\RequirePackage{fix-cm}
\documentclass{svjour3}                     % onecolumn (standard format)
\smartqed  % flush right qed marks, e.g. at end of proof
\usepackage{graphicx}
\usepackage{amssymb}
\usepackage{graphicx}
\usepackage{amsmath}
\usepackage{multirow,tabularx}
\usepackage{hhline}
\usepackage{enumitem}
\usepackage{subfig}
\usepackage{url}
\usepackage{amssymb}
\usepackage{url}
\usepackage{ragged2e} 
\usepackage{adjustbox}

\usepackage{setspace}
\usepackage{rotating}

\usepackage[toc,page]{appendix}

\usepackage{pdfpages}
\usepackage{booktabs}

\usepackage{caption}
\begin{document}

\title{Joint Surgical Gesture and Task Classification with Multi-Task and Multimodal Learning %\thanks{Grants or other notes
}
%\subtitle{ Learning Shared and Task-Specific Representations\\ If so, write it here}


\author{Duygu Sarikaya  \and  Khurshid A. Guru \and  Jason J. Corso  %etc.
}


\institute{D. Sarikaya \at
              Department of Computer Science
and Engineering, SUNY Buffalo, NY 14260-1660 USA \\
              \email{duygusar@buffalo.edu }           %  \\
           \and
           K. A. Guru \at
             Applied Technology Laboratory for Advanced
             Surgery, Roswell Park Cancer Institute, Buffalo, NY 14263 USA
               \and
            J. J. Corso \at Department of Electrical Engineering and
        Computer Science, University of Michigan, Ann Arbor, MI 48109 USA
}

\date{Received: date / Accepted: date}

\maketitle

\begin{abstract} 
~\\
~\\
\textbf{Purpose}~\\
We propose a novel multi-modal and multi-task architecture for simultaneous low
level gesture and surgical task classification in Robot Assisted Surgery (RAS)
videos. 
~\\
~\\
\textbf{Methods}~\\
Our end-to-end architecture is based on the principles of a long short-term memory
network (LSTM) that jointly learns temporal dynamics on rich representations of
visual and motion features, while simultaneously classifying activities of low-level
gestures and surgical tasks. 
~\\
~\\
\textbf{Results}~\\
Our experimental results show that our approach is superior compared to an architecture that classifies the gestures and surgical tasks separately on visual cues and motion cues respectively. We train our model on a fixed random set of 1200 gesture video segments and use the rest 422 for testing. This results in around 42,000 gesture frames sampled for training and 14,500 for testing. For a 6 split experimentation, while the conventional approach reaches an Average Precision (AP) of only 29\% (29.13\%), our architecture reaches an AP of 51\% (50.83\%) for 3 tasks and 14 possible gesture labels, resulting in an improvement of 22\% (21.7\%).
~\\
~\\
\textbf{Conclusions}~\\
Our architecture learns temporal dynamics on rich representations of visual and
motion features that compliment each other for classification of low-level gestures
and surgical tasks. Its multi-task learning nature makes use of learned joint relationships and combinations of shared and task-specific representations. While
benchmark studies focus on recognizing gestures that take place under specific
tasks, we focus on recognizing common gestures that reoccur across different tasks
and settings and significantly perform better compared to conventional architectures.
\keywords{Robot-Assisted Surgery \and Surgical Gesture Classification  \and Multi-task Learning \and Multimodal Learning  \and Long Short-term Recurrent Neural Networks \and Convolutional Neural Networks}
\end{abstract}

\section{Introduction}

Video understanding of robot-assisted surgery (RAS) videos is an active research area. Modeling the gestures and skill level of surgeons presents an interesting problem for the needs addressed by the community such as automation and early identification of technical competence for surgeons in training. We approach the problem of video understanding of RAS videos as modeling the motions of surgeons. By analyzing the scene and object features, motion, low-level surgical gestures and the transitions among the gestures, we could model the activities taking place during surgical tasks \cite{lingling}. The insights drawn may be applied in effective skill acquisition, objective skill assessment, real-time feedback, and human-robot collaborative surgeries \cite{duygu}.

We propose to model the low-level surgical gestures; recurring common activity segments as described by Gao \textit{et al.} \cite{jigsaws} and the surgical tasks that are composed of gestures with a multimodal and multi-task learning approach. We use the different modalities of the visual features and motion cues (optical flow), and learn the temporal dynamics jointly. We argue that surgical tasks are better modeled by the visual features that are determined by the objects in the scene, while the low-level gestures are better modeled with motion cues. For example, the gesture \textit{Positioning the needle} might take place in different tasks of \textit{Suturing} and \textit{Needle Passing} \cite{jigsaws}. Or, on a higher level of actions, placing a \textit{Tie Knot} might occur during a task of suturing on a tissue and a more specific and challenging task of Urethrovesical Anastomosis (UVA) \cite{duygu} that involves stitching and reconnecting two parts together. If we rely on visual features of the object and the scenes only, these smaller segments of activities would have very different representations. Motion cues of low-level gestures, on the other hand, are independent of  object and scene features. This helps us to classify common low-level gestures that are generic in nature and that reoccur in different tasks with different objects and under different settings better, and also helps reduce overfitting. However, we also believe that visual features are complementary to the motion cues, which are independent of object and scene features. The gesture ``Positioning the needle" might take place in different tasks of \textit{Suturing} and \textit{Needle Passing}, however is not likely to take place in \textit{Knot Tying} tasks as a needle is not used for this task. 

\begin{figure}[!t] 
\centering
 {\includegraphics[width = 4in]{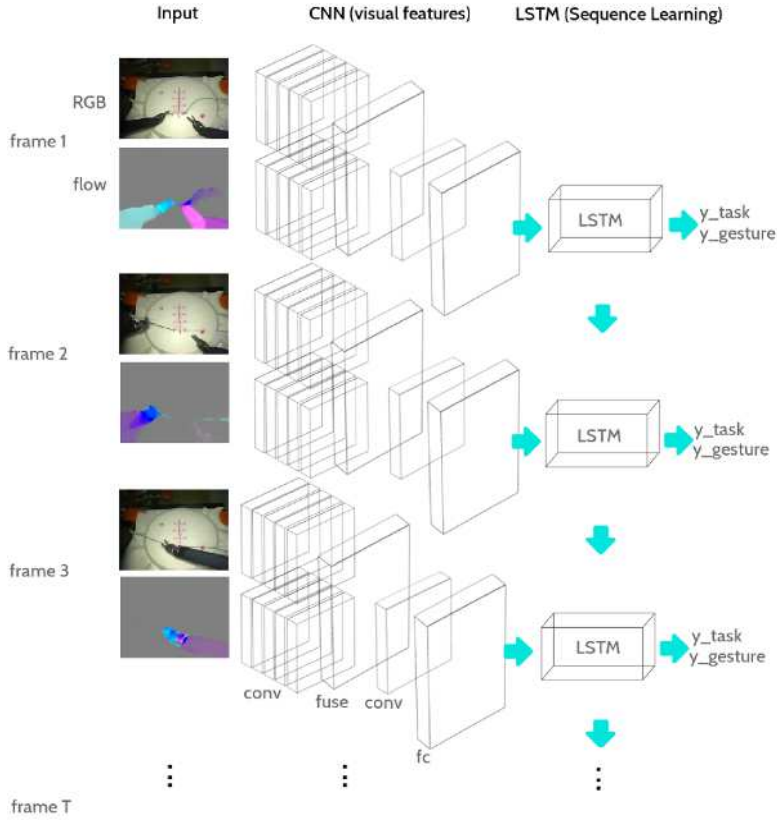}}
\caption{We propose a novel architecture that is based on the principles of LRCN, a specialized LSTM network that jointly learns temporal dynamics and visual features by convolutional network models. Our architecture however, jointly learns on the two modalities of the video:  RGB frames and the RGB representation of the optical flow information. We extract higher level features using CNNs that we train on separate tasks. After we convolve the two streams of input pairs, we concatenate the CNN features and use it as an input to our recurrent joint model that learns temporal dynamics of consequent frames. Our architecture simultaneously classifies the common low-level gestures and the surgical tasks.}
\label{fig:1}
\end{figure}

In this paper, we address the problem of simultaneous classification of common low-level gestures and surgical tasks. While surgical task recognition, which is characteristically defined by the scene and the objects, is an easy task, the low-level gesture recognition remains a challenge. In our work, we focus on this challenging task by making use of complex relationships between the visual and motion cues, and the relationship between surgical activities at different levels of complexity. We propose a novel architecture that simultaneously classifies the common low-level gestures and the surgical tasks by making use of these joint relationships, combining shared and task-specific representations with a multi-task learning approach. Moreover, our architecture supports multimodal learning and is trained on both visual features and motion cues to achieve better performance. An overview of our method is shown in Figure \ref{fig:1}. We use the inputs of RGB frames for visual features and the RGB representation of the optical flow information of the same frames to refer to motion cues. We extract the high level features of these inputs using the convolutional neural networks we have trained on each task separately and use them as input to our recurrent joint model. After we convolve the two streams of input modality pairs, we concatenate the higher level features and use it as an input to our recurrent model that learns temporal dynamics of consequent frames.

\section{Related Work}

Surgical activity recognition is an active research area that gets a lot of attention from the computer vision and medical imaging communities. The potential of autonomous or human-robot collaborative surgeries, automated real-time feedback, guidance and navigation during surgical tasks is exciting for the community. There are studies that address this open research problem. The methods range from using SVM classification with Linear Discriminant Analysis (LDA) \cite{lin2005,lin2006}, Gaussian mixture models (GMMs) \cite{leong},  variations of Hidden Markov Models (HMM) that represents probability distributions over sequences of observations, \cite{yang,varadarajan} and more recently, to convolutional neural networks and recurrent neural networks \cite{m2cai_smooth,colin,m2cai_lstm}. 

Ahmidi \textit{et al.} \cite{jigsaws_benchmark} do a comparative benchmark study on the recognition of gestures on JIGSAWS dataset.  In this study, in order to classify surgical gestures, three main methods are chosen:  Bag of Spatio-Temporal Features (BoF), Linear Dynamical System (LDS) and a composite Gaussian Mixture Model- Hidden Markov Model: GMM-HMM.  HMMs are often used in studies that use additional modalities of various sensors \cite{sensor1,sensor2}, and to classify high level surgical activities \cite{sensor3}. 

Deep neural networks have been introduced to the activity recognition tasks \cite{simonyan,lrcn}. However, in the medical field these advances are only very recently being explored. DiPietro \textit{et al.} \cite{colin} propose using Recurrent Neural Networks (RNN) trained on kinematic data for surgical gesture classification on JIGSAWS dataset. Some other recent studies focus on classifying surgery phases; a higher level of surgical activity that includes a sequence of different tasks. A recent work by Cad\'ene \textit{et al}. \cite{m2cai_smooth} uses deep residual networks to extract visual features of the video frames and then applies temporal smoothing with averaging. The authors of this work finally model the transitions between the surgery phase steps with an HMM. Twinanda \textit{et al.} \cite{m2cai_lstm} offer a study on classification of surgery phases by first extracting visual features of video frames via a CNN and then passing them to an SVM to compute confidences of a frame belonging to a surgery phase. These confidences are used as inputs to an HMM and also to an LSTM network. 

HMMs are often used in surgical activity recognition as they address the temporal dynamics of the gestures and relationships between surgical phases. Although there are studies that rely on visual features of the surgical video frames,  a large number of promising studies rely highly on kinematic data. The problem with capturing kinematic data is that; it requires additional equipment, and it is possible that the kinematic data is sampled over periods and might be missing important motion information. For the latter case, we need to preprocess the kinematic data by interpolation to estimate the motion information of these missing parts. Kinematic data also requires preprocessing based on specific attributes by the equipment that has captured it. Visual features on the other hand, when used alone, are highly dependent on the objects and scenes, and might be missing important motion cues.

In this paper, we address the video understanding problem of activity classification in Robot-assisted Surgery (RAS) videos. We simultaneously classify common low-level gestures and surgical tasks. We focus on modeling the common low-level gestures that are generic and that reoccur in different tasks. We believe that the low-level gestures are better modeled by motion cues, while the surgical tasks are better modeled by the visual features that are characteristically defined by the objects in the scene. However, we also believe that visual features are complementary to the motion cues, which are independent of object and scene features.

Many recent studies have shown that exploiting the relationship across different tasks, jointly reasoning multi-tasks \cite{ubernet,mingsheng,yao} and taking advantage of a combination of shared and task-specific representations \cite{crossstitch} perform astonishingly better than their single-task counterparts. 

We propose using recurrent neural networks \cite{rnn}, specifically long term short memory (LSTM) neural networks \cite{lstm}, in order to model the complex temporal dynamics and sequences. We aspire to model the sequences of gestures occurring in surgical-task videos with a model that is deep over temporal dimensions. Our architecture is based on the principles of LRCN \cite{lrcn}, a specialized LSTM network that jointly learns temporal dynamics and visual features by convolutional network models. Our model supports multimodal learning, and uses the RGB frames and the RGB representation of the optical flow information relating the motion cues in the same frames as input. Our architecture simultaneously classifies the low-level gestures and surgical tasks by a multi-task learning approach making use of joint relationships, combining shared and task-specific representations. In this sense, our model is truly end-to-end and novel in the way that it supports multimodal and multi-task learning for surgical low-level gesture classification.

\begin{figure}[!t]
\centering
\subfloat{\includegraphics[width = 1in]{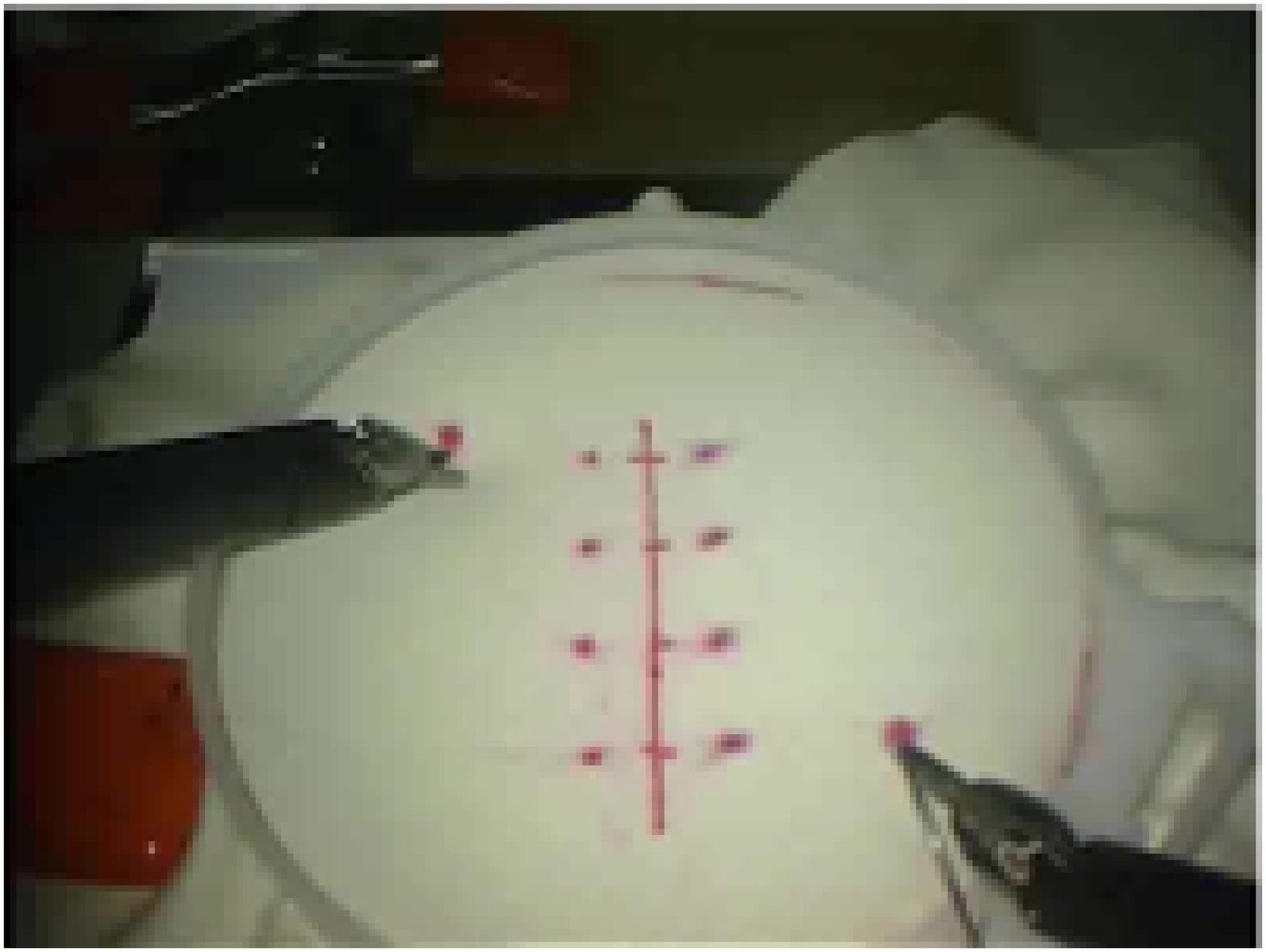}\includegraphics[width = 1in]{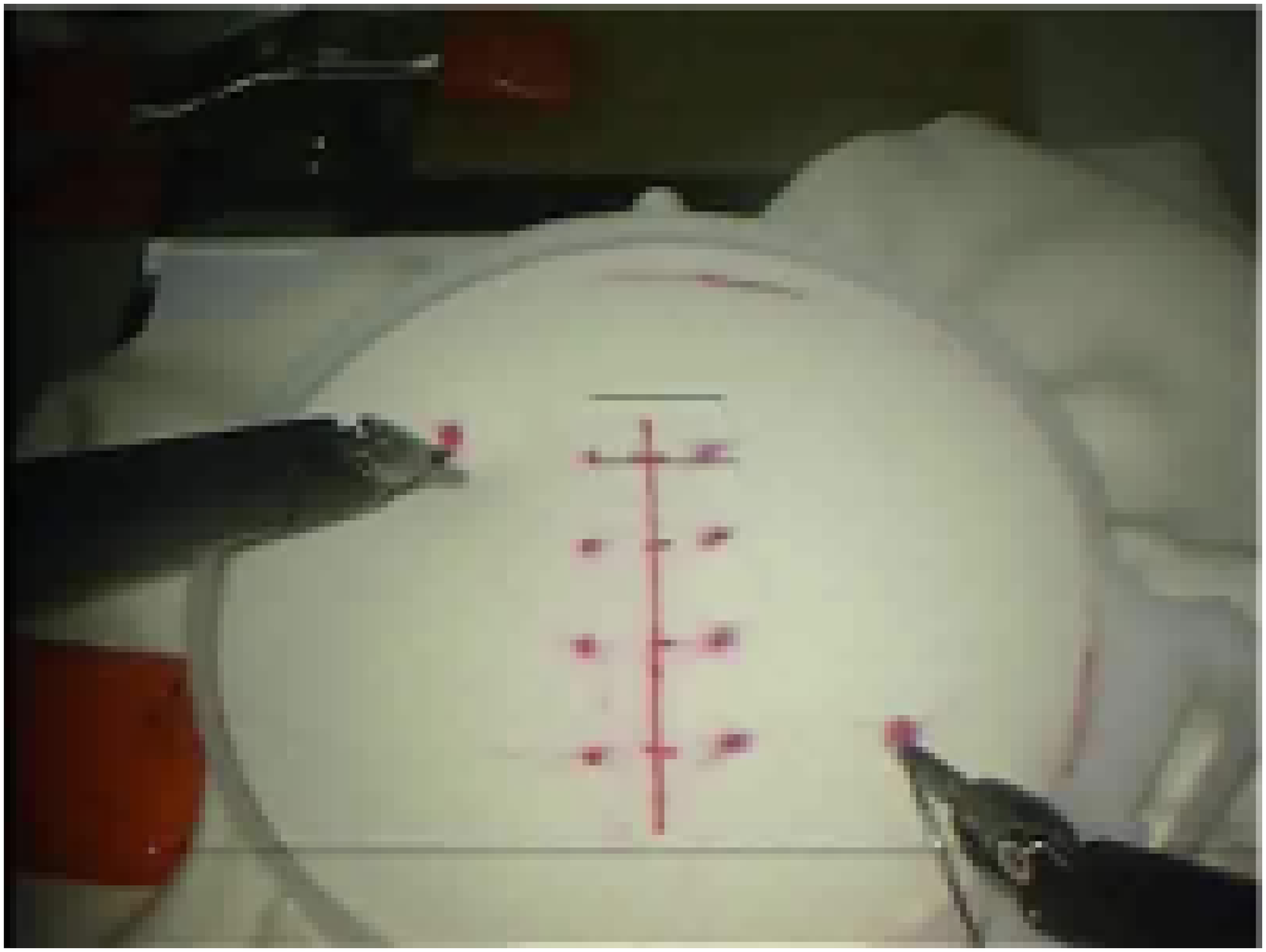}\includegraphics[width = 1in]{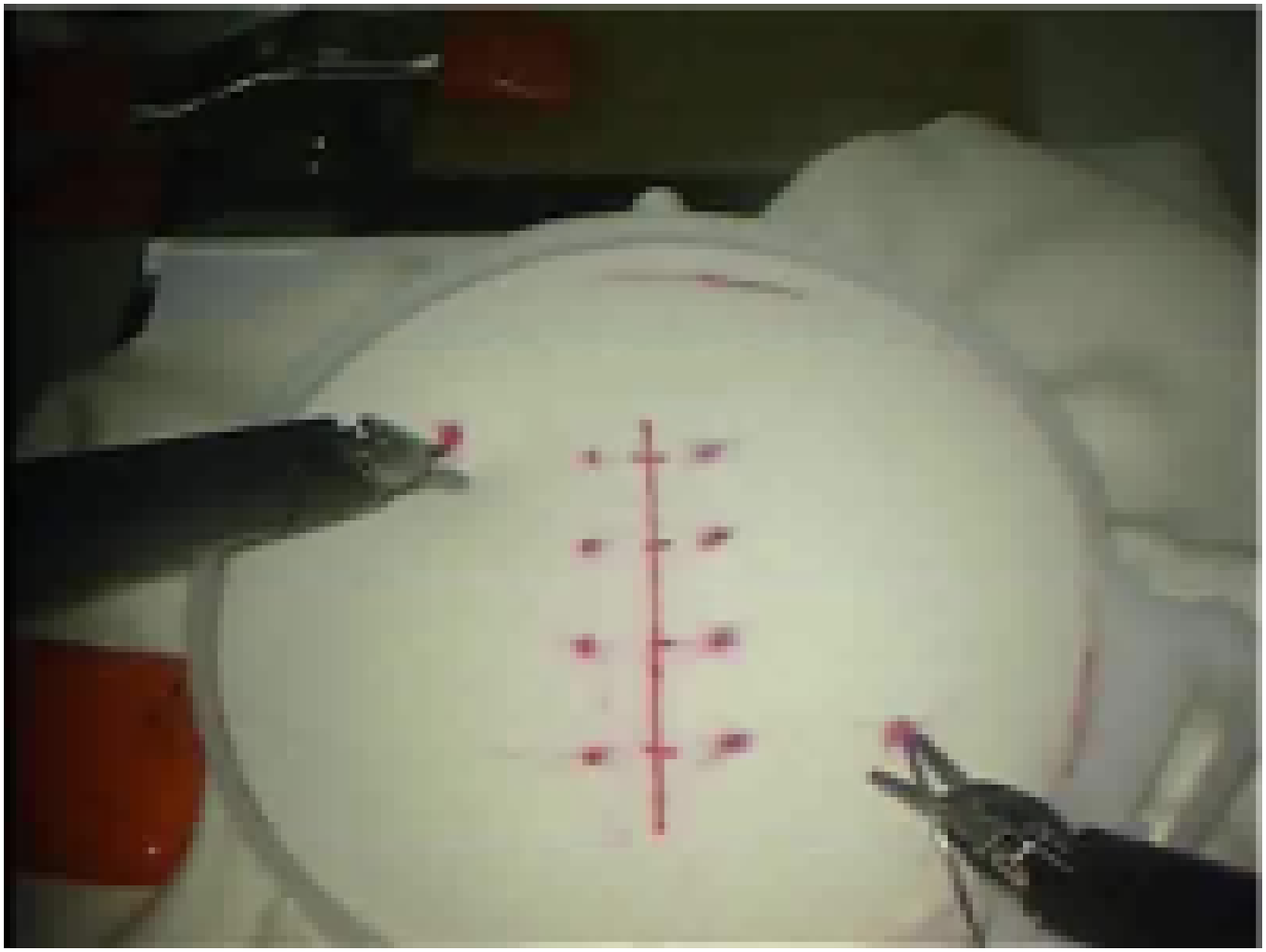}\includegraphics[width = 1in]{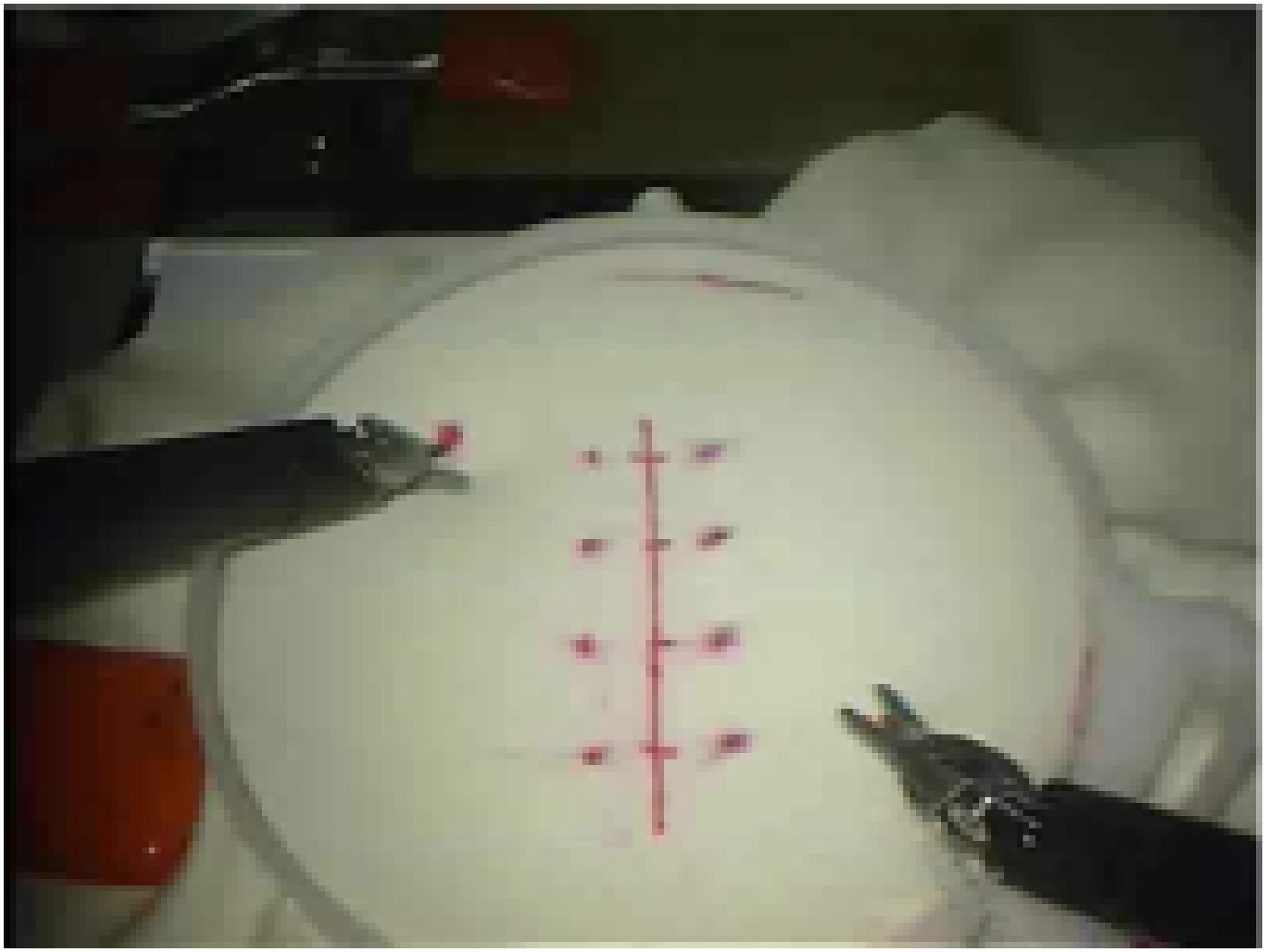}} \\
\includegraphics[width = 1in]{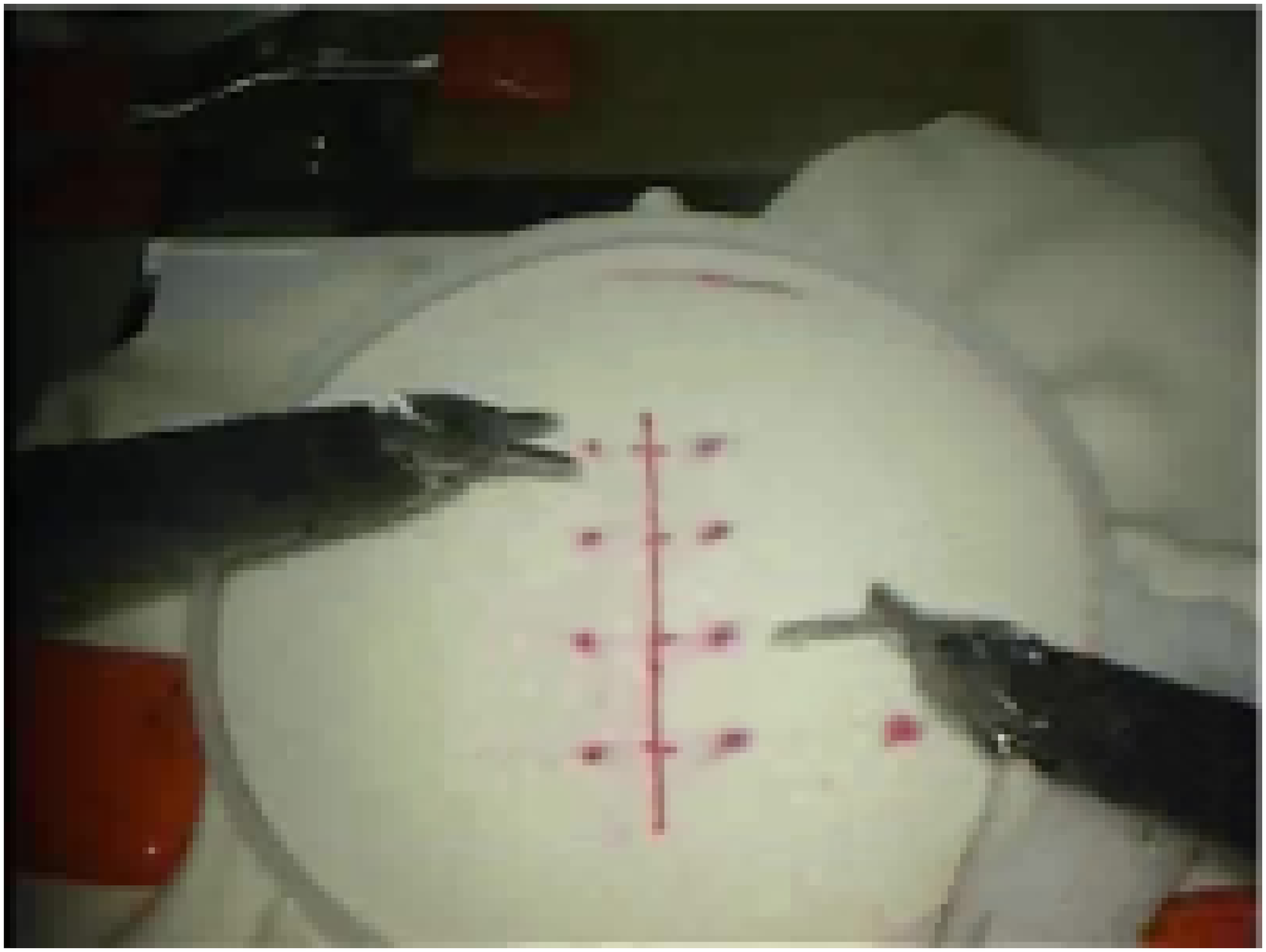}\includegraphics[width = 1in]{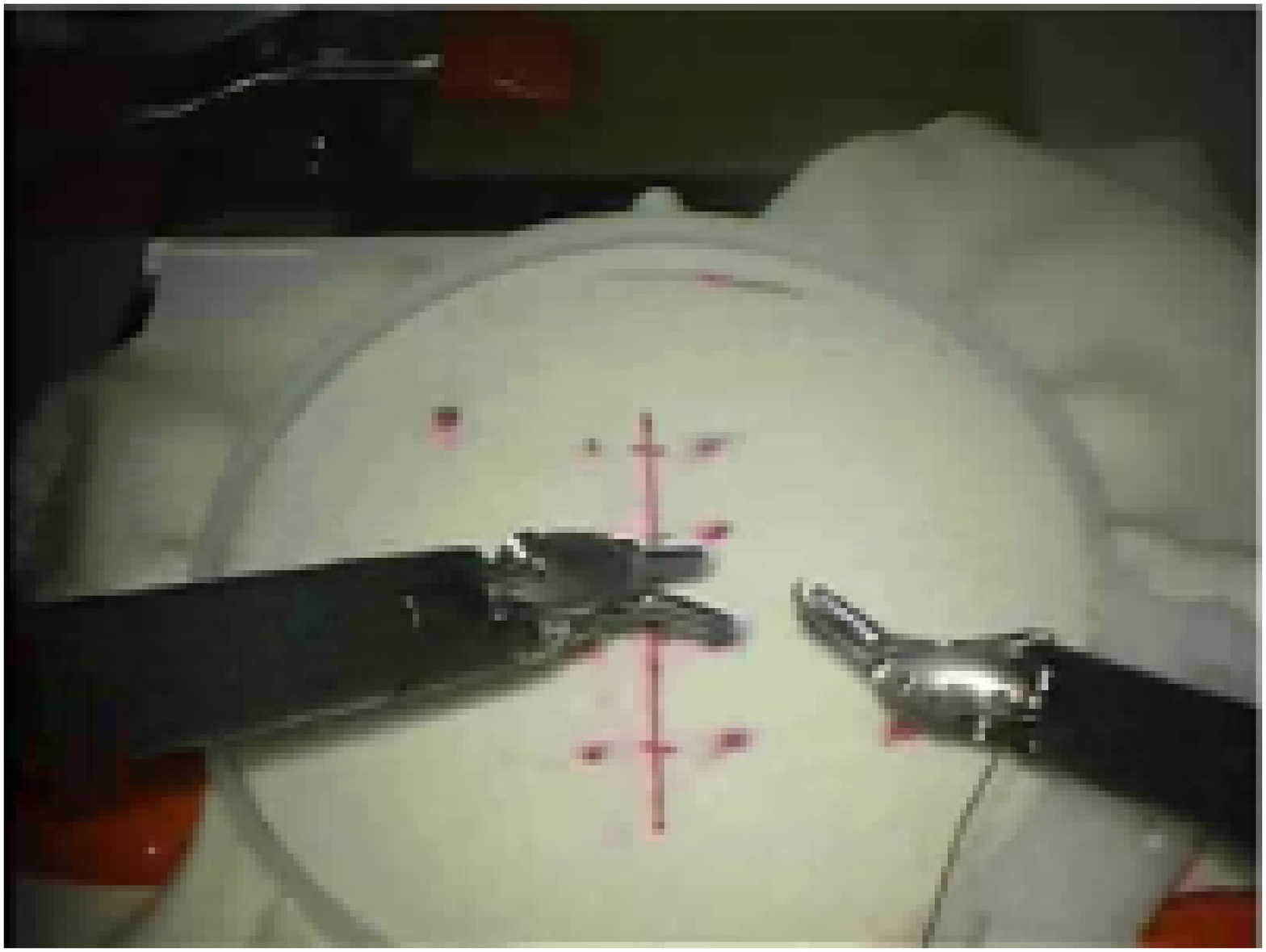}\includegraphics[width = 1in]{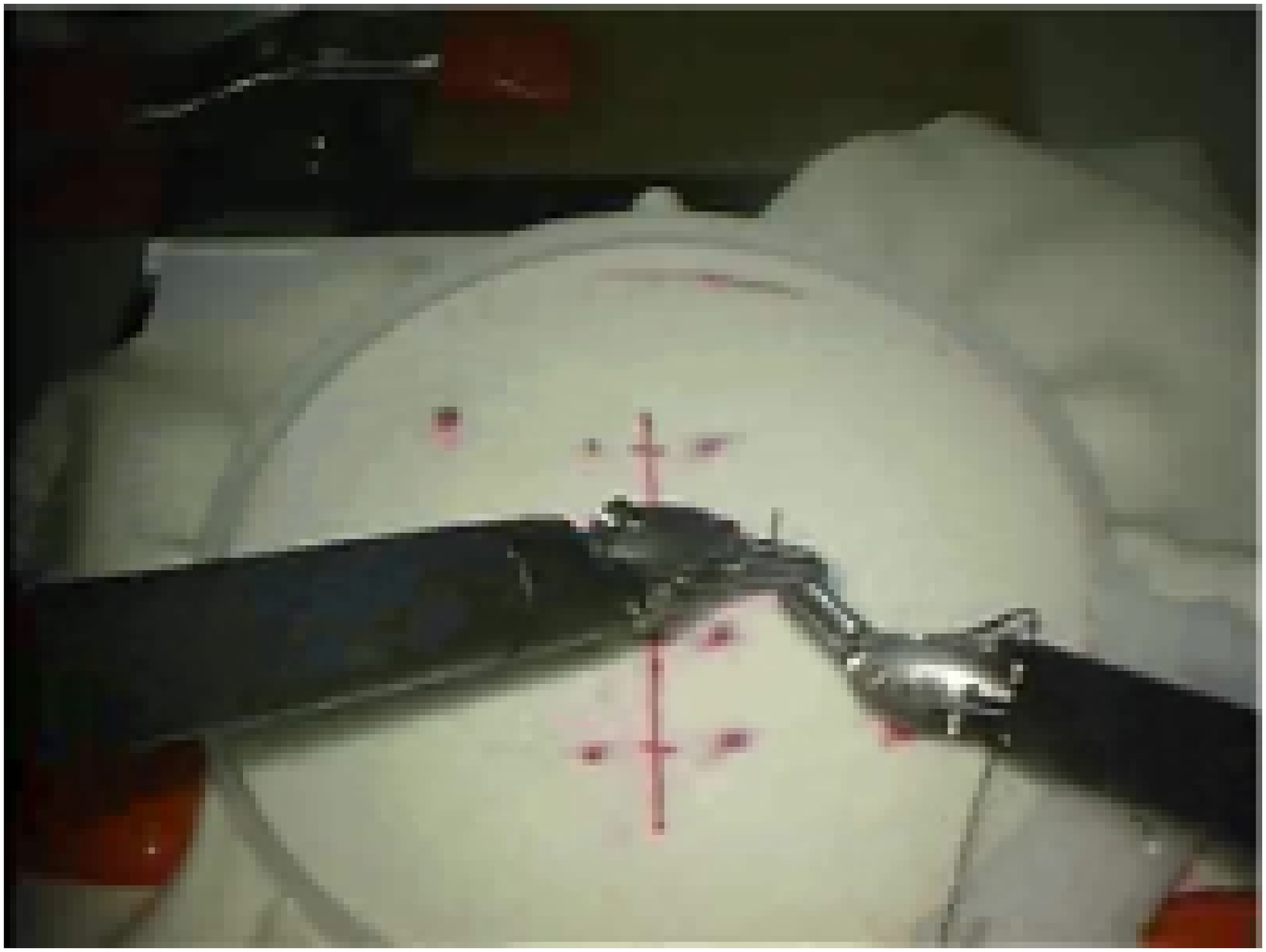}\includegraphics[width = 1in]{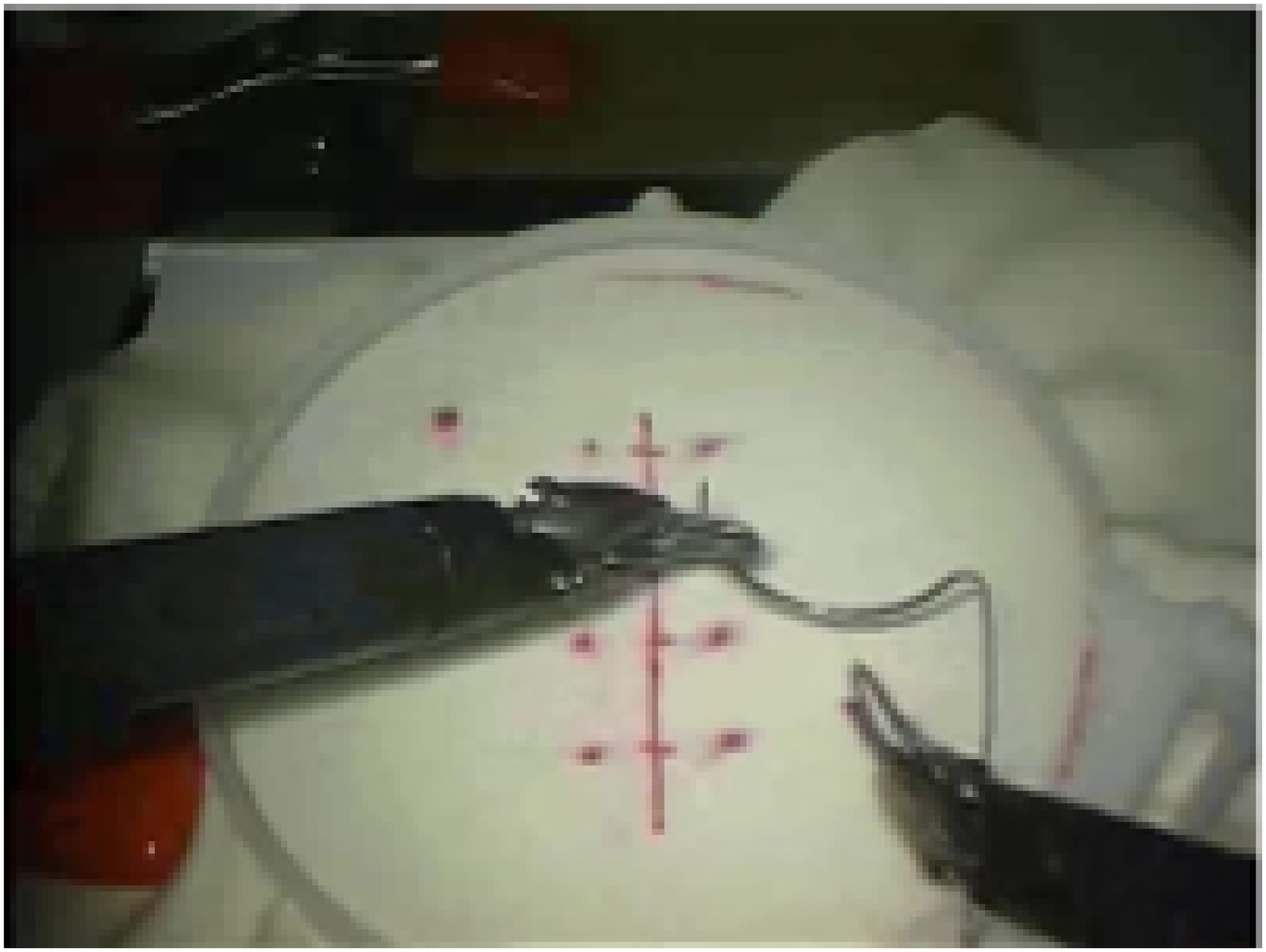} \\
\includegraphics[width = 1in]{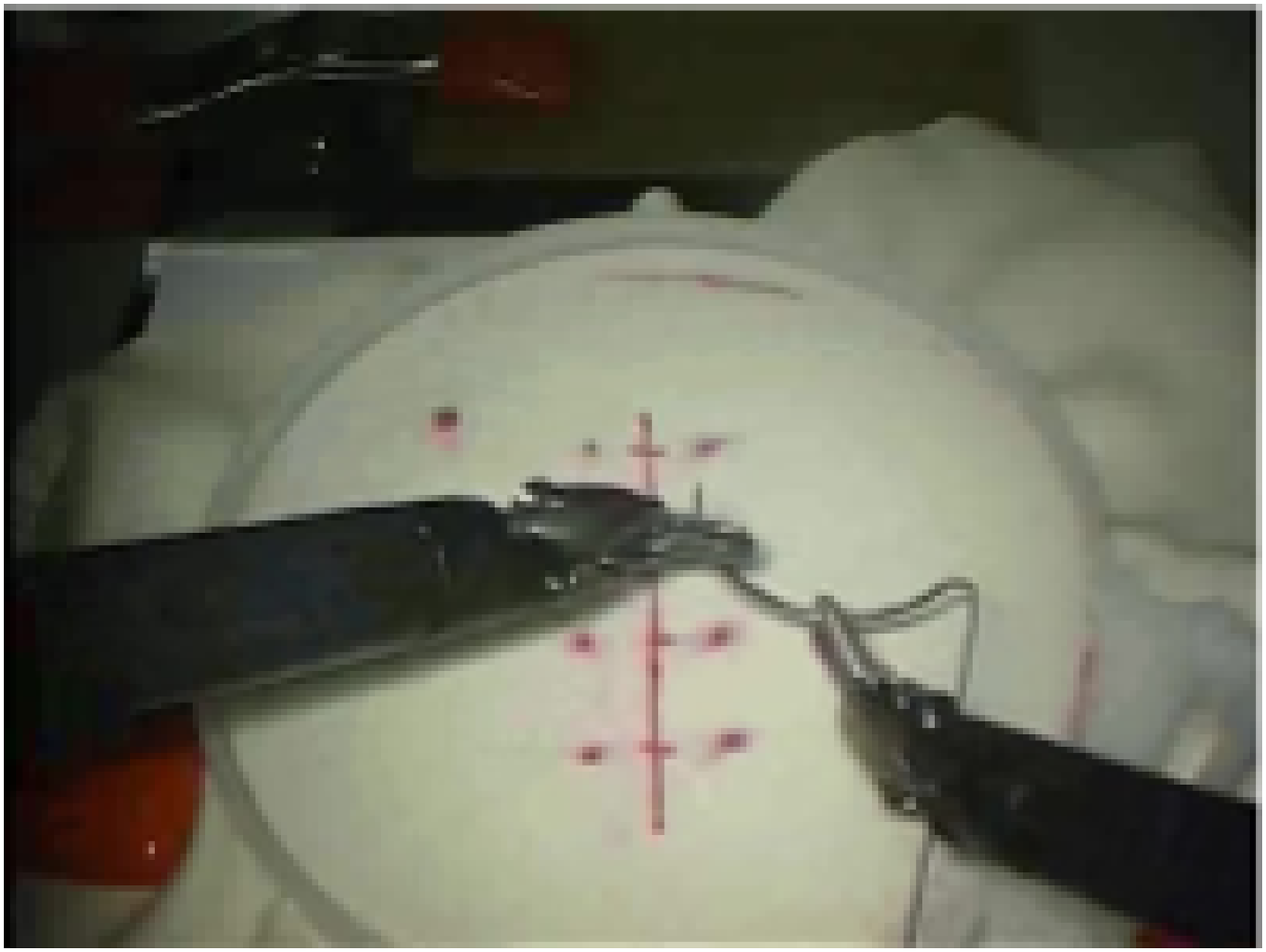}\includegraphics[width = 1in]{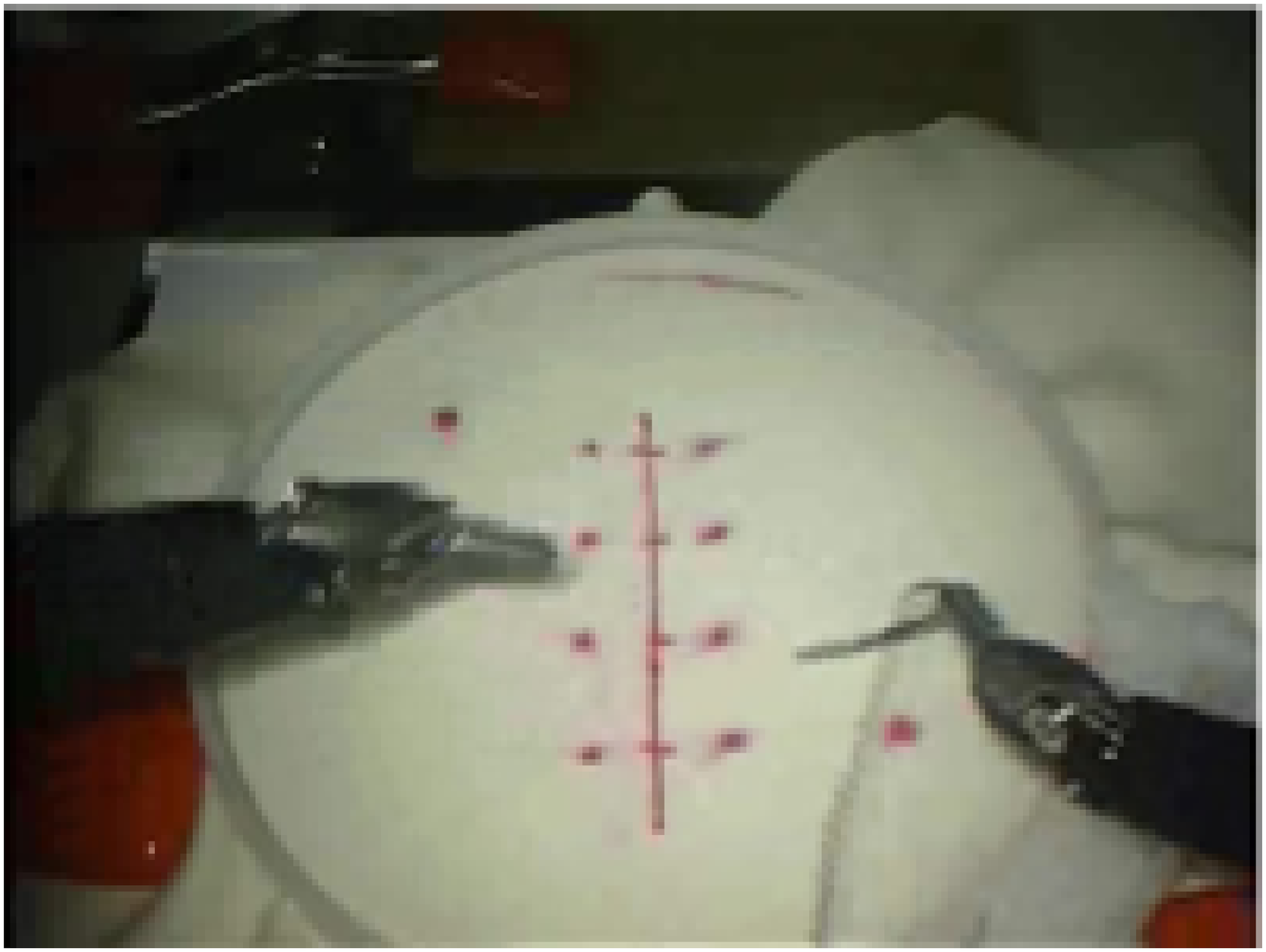}\includegraphics[width = 1in]{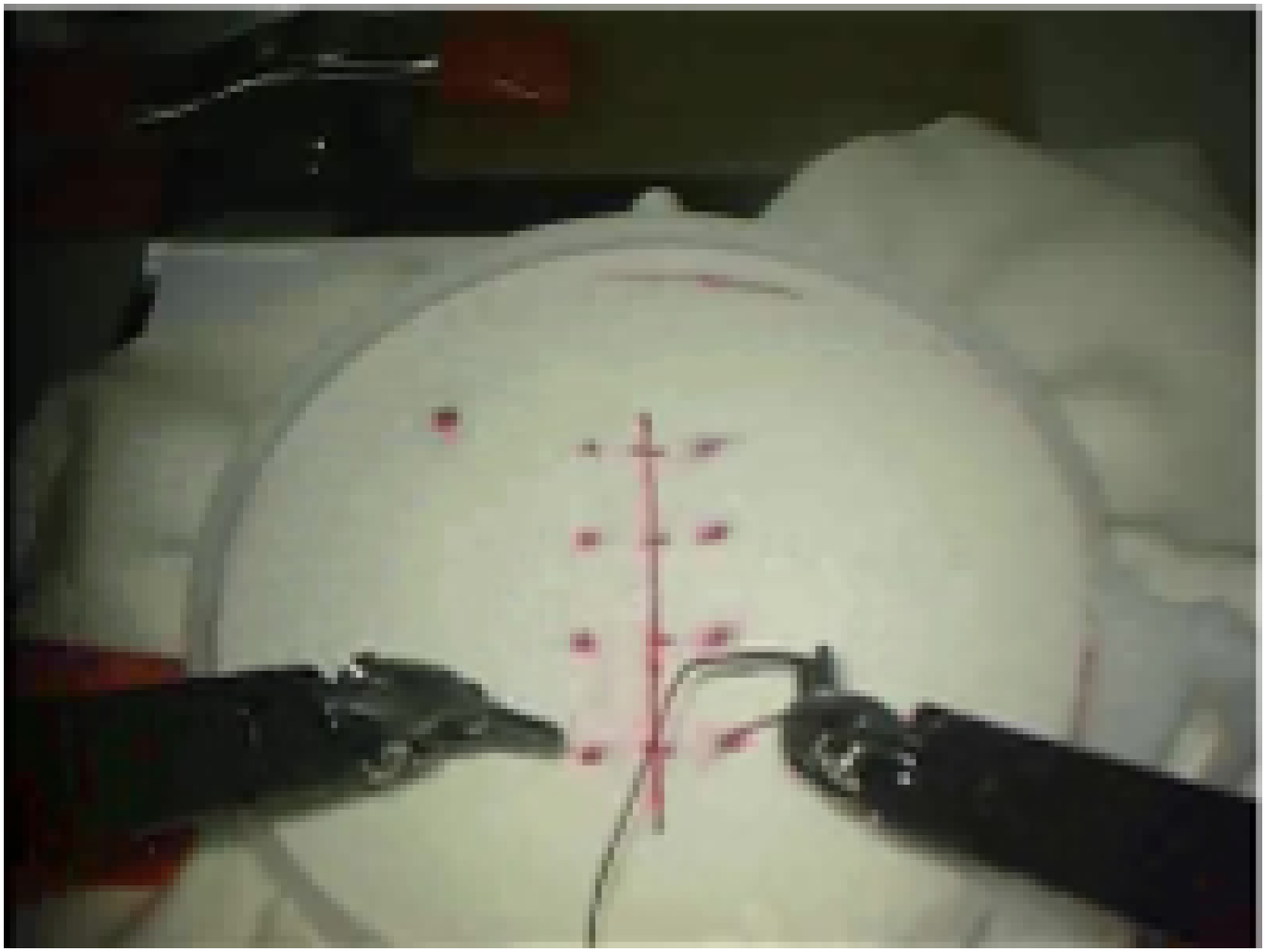}\includegraphics[width = 1in]{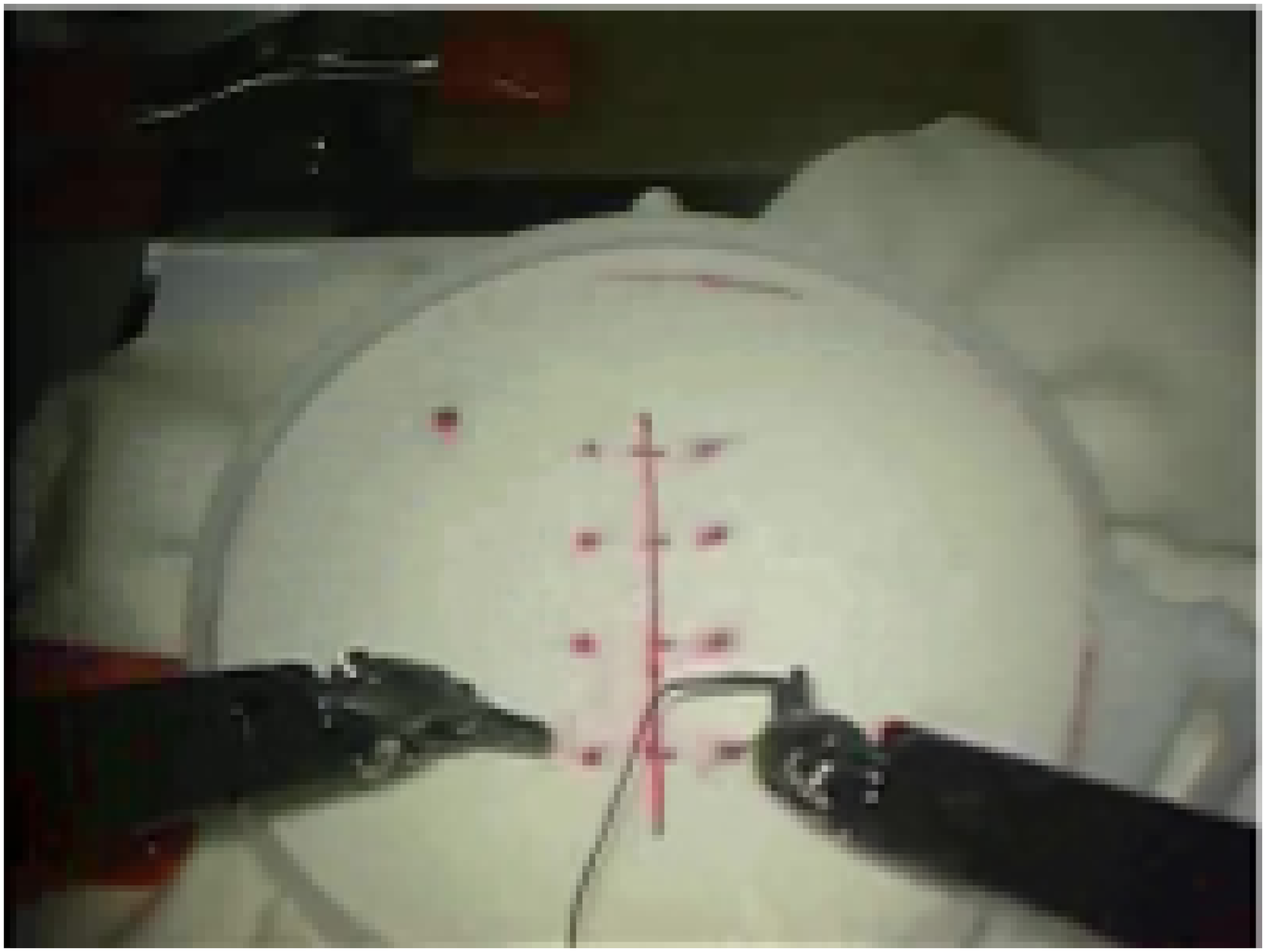} \\
\includegraphics[width = 1in]{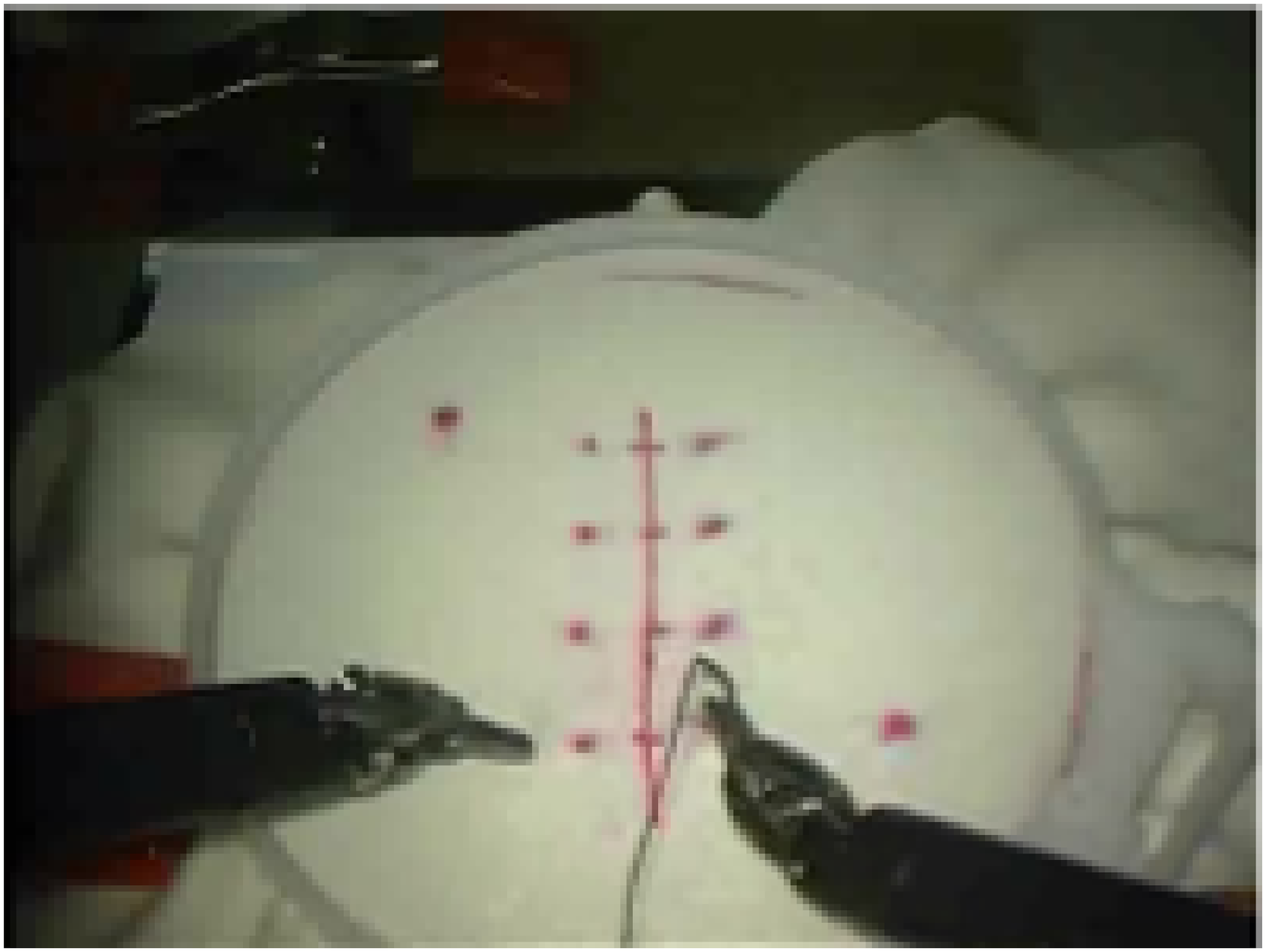}\includegraphics[width = 1in]{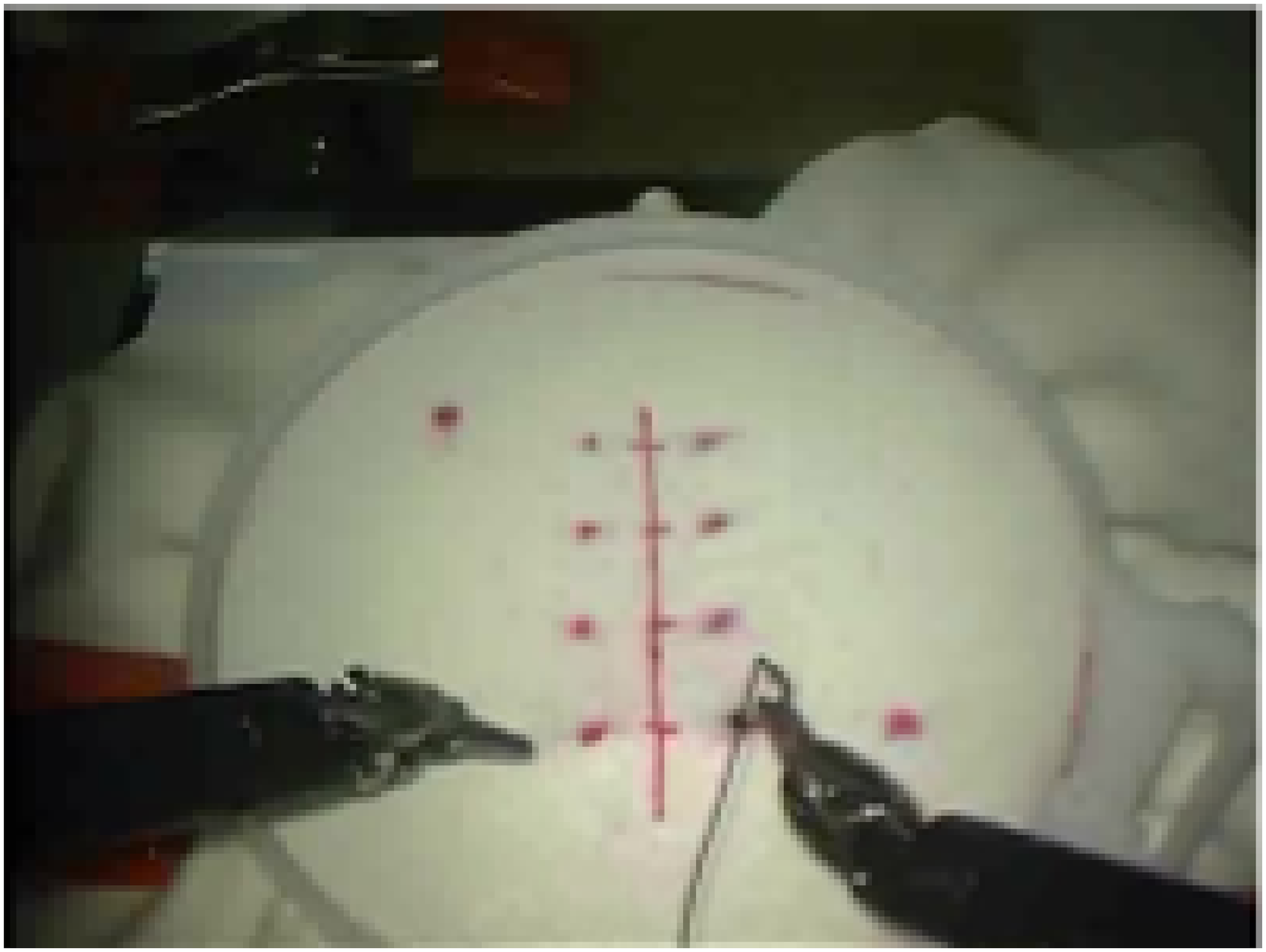}\includegraphics[width = 1in]{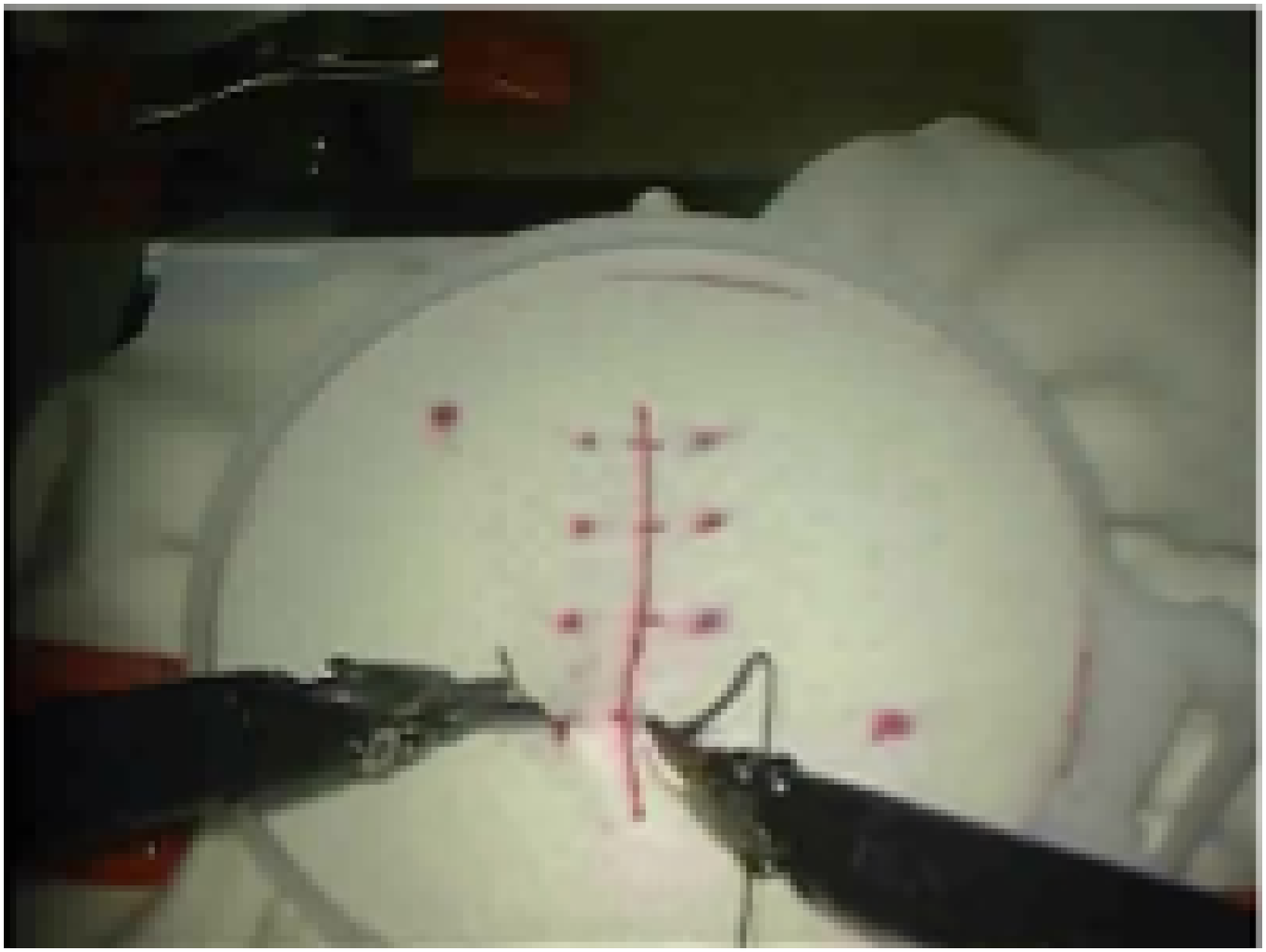}\includegraphics[width = 1in]{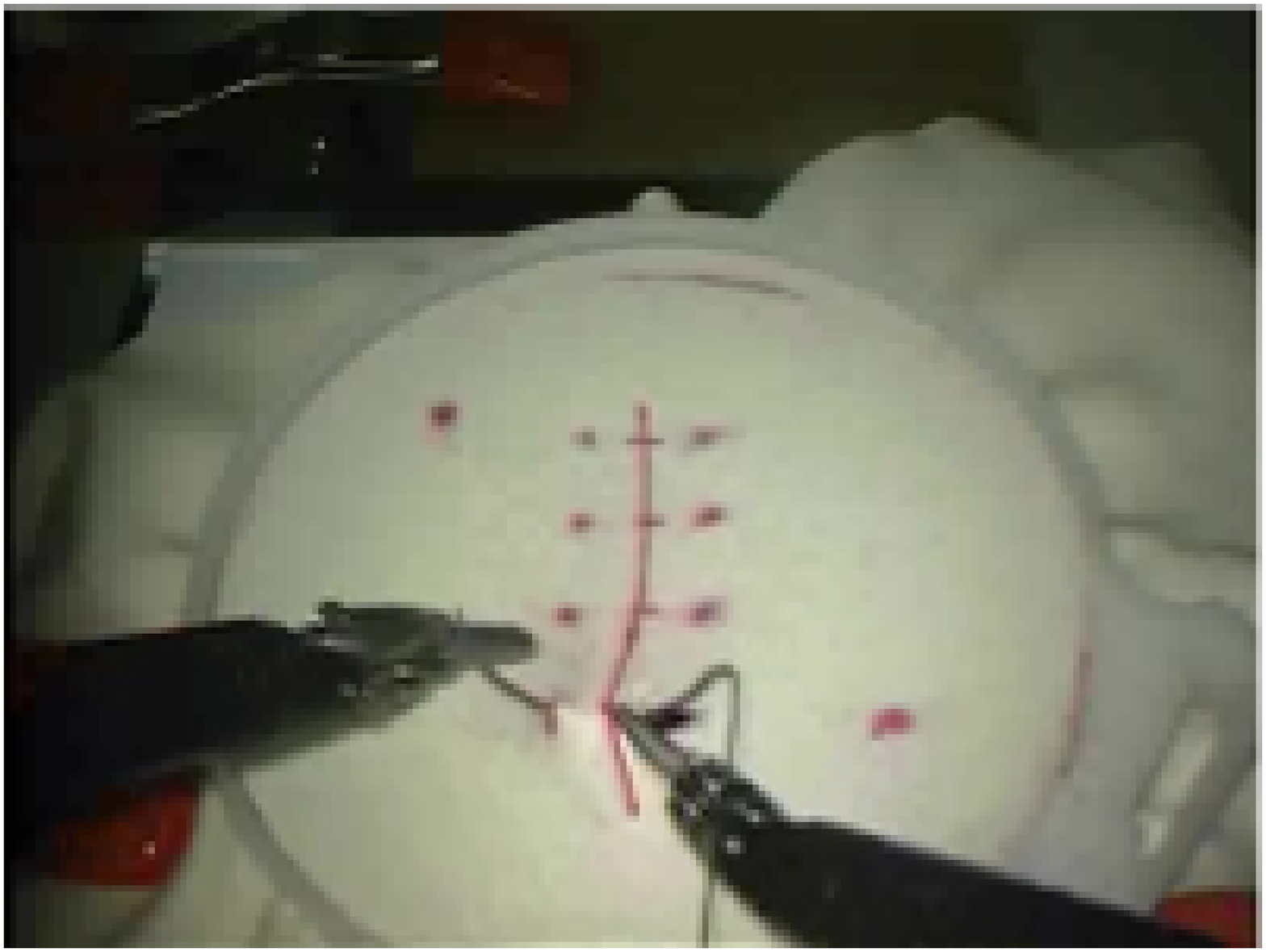} \\
\includegraphics[width = 1in]{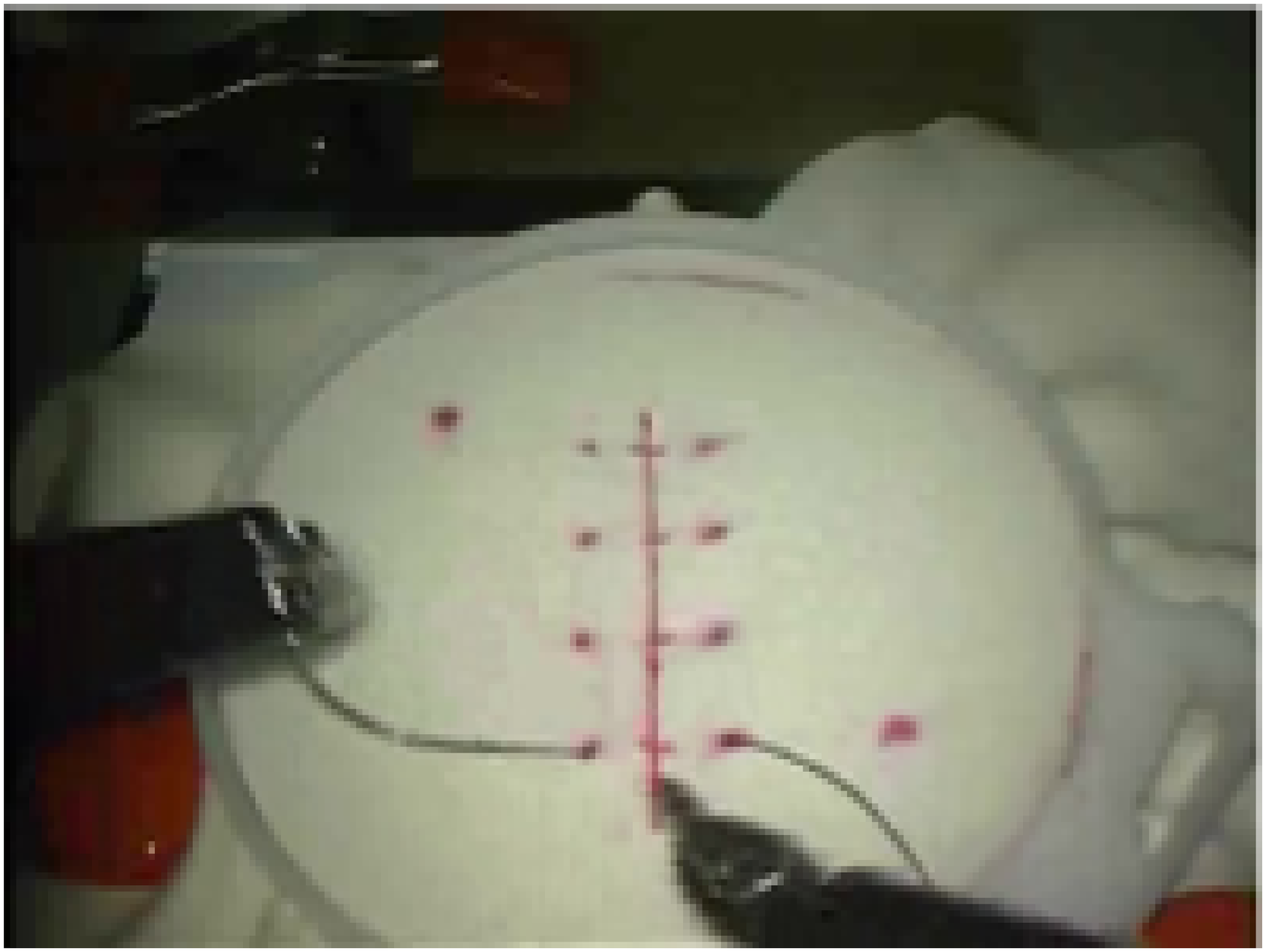}\includegraphics[width = 1in]{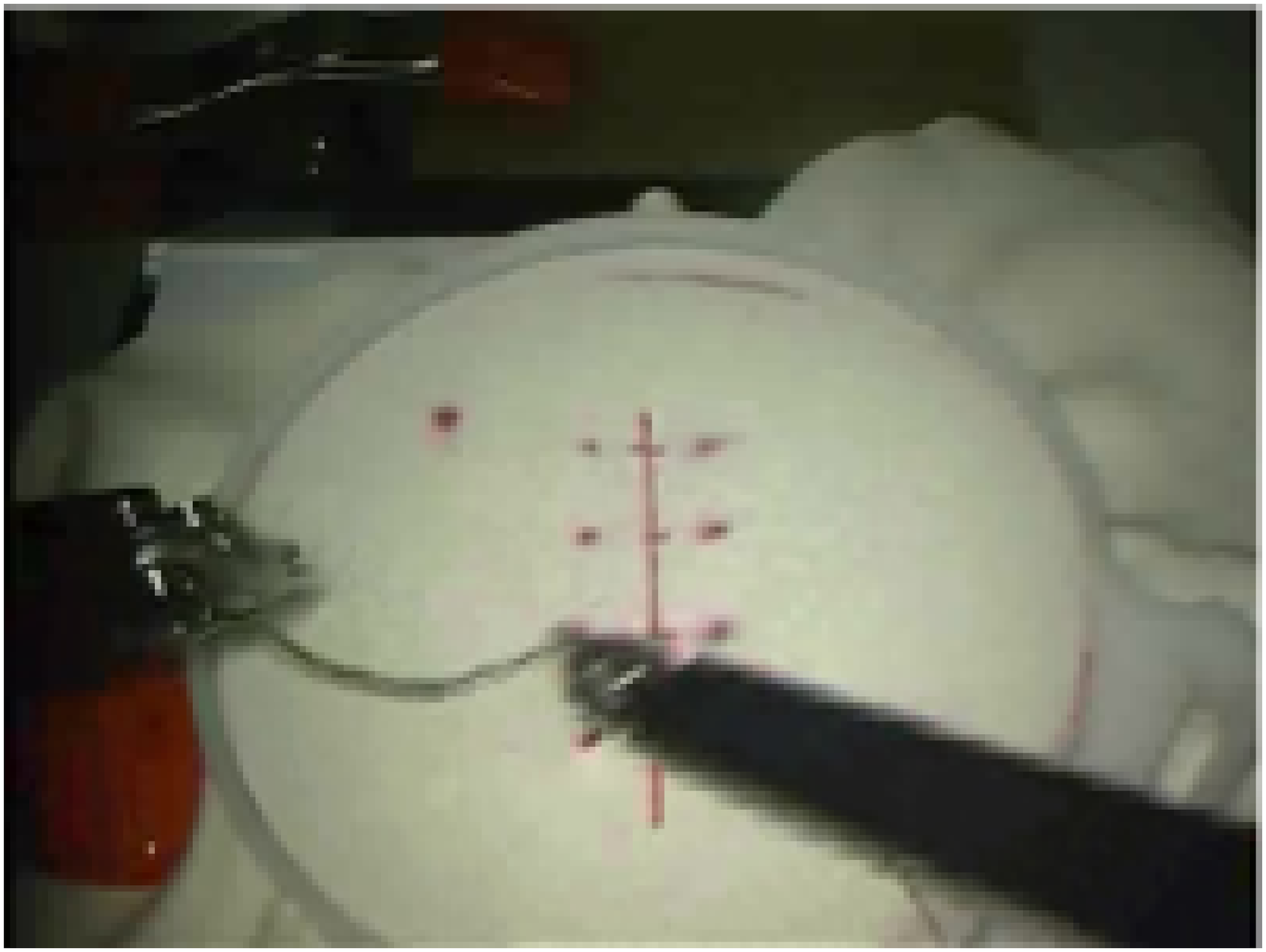}\includegraphics[width = 1in]{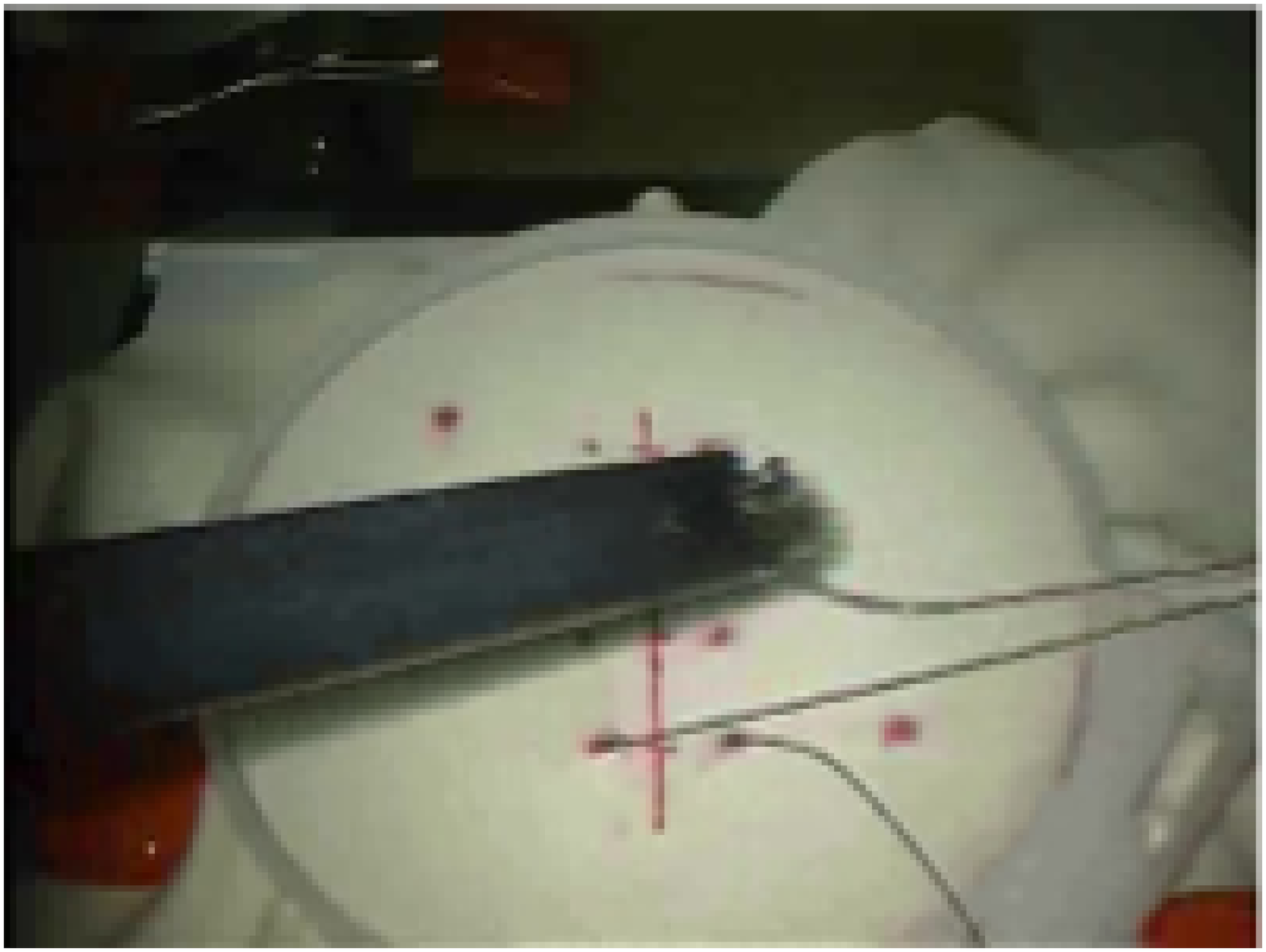}\includegraphics[width = 1in]{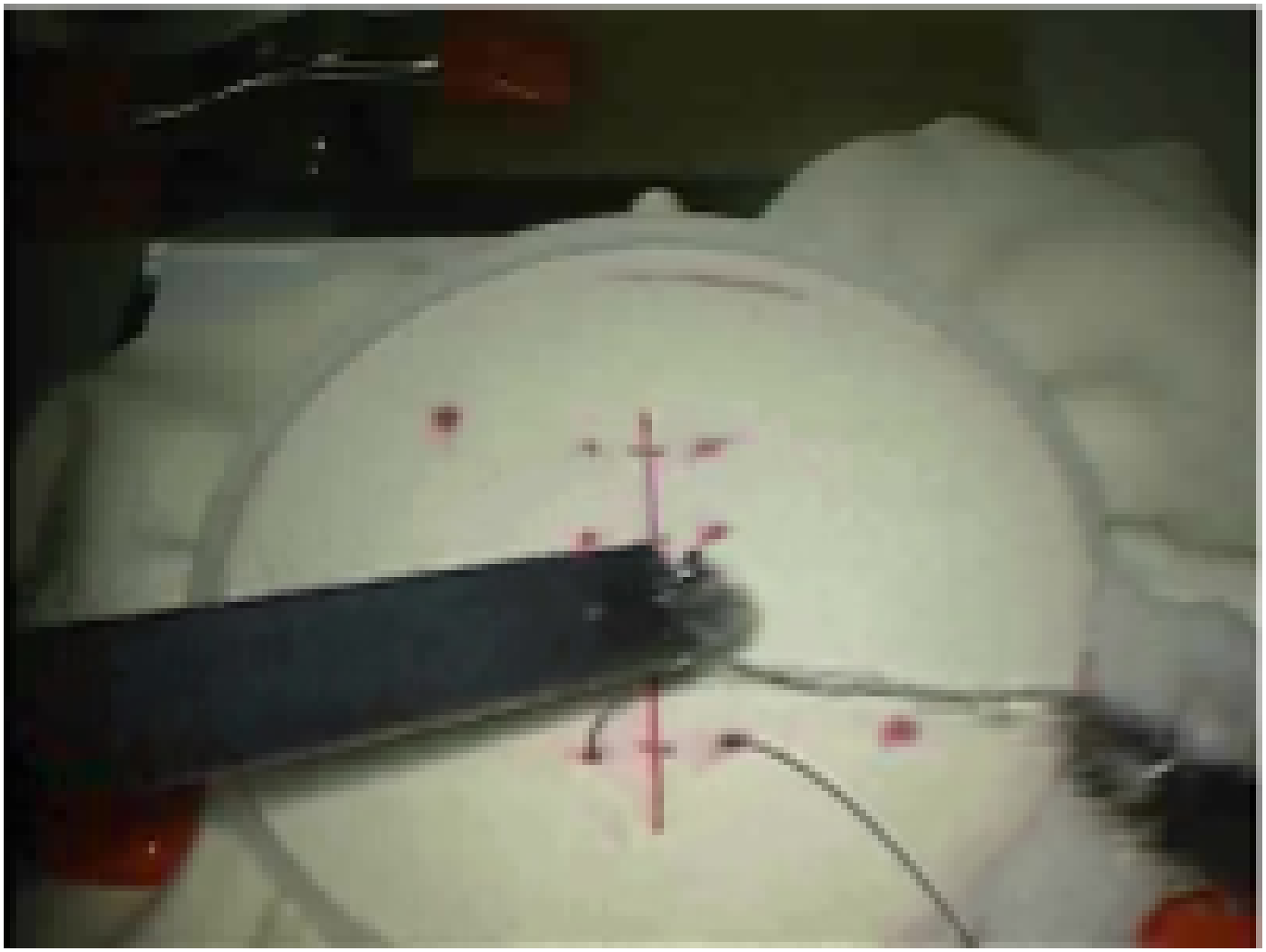}
\caption{ A temporal sequence of surgical activities during a \textit{Suturing} task are shown (from left to right, top to bottom). The surgeon does a suture on the tissue following the guide landmarks. }
 \label{fig:2}
\end{figure}

\section{Dataset} The JHU-ISI Gesture and Skill Assessment Working Set (JIGSAWS) \cite{jigsaws} provides a public benchmark surgical activity dataset. In this video dataset, $8$ surgeons with varying expertise perform $3$ surgical tasks on the daVinci Surgical System (dVSS $^{\tiny{\textregistered}}$)  (Figure \ref{fig:2}). The dataset includes video data recorded during the performance of the tasks: \textit{Suturing}, \textit{Needle Passing} and \textit{Knot Tying} and provide $15$ low-level gesture labels, which are the smallest action units where the movement is intentional and is carried on towards achieving a specific goal. The gestures reoccur in various tasks. A surgical gesture is a small, however a meaningful and purposeful segment of activity that, the tasks could be broken into. For example, the gesture of \textit{G2: Positioning needle} could take place in both \textit{Suturing} and \textit{Needle Passing} tasks. However, the objects and the scene in these two tasks are quite different; while \textit{Suturing} involves a task of placing sutures on a tissue model, \textit{Needle Passing} takes place on a set of metal hoops \cite{jigsaws}. The gestures form a common vocabulary for small action segments that reoccur in different tasks. We focus on recognizing these low-level gestures, even when they reoccur across different tasks with varying accompanying objects and are performed on different sets.

\begin{table}[]
\centering
\begin{adjustbox}{width = 4in }
\begin{tabular}{|l|l|}
\hline
\multicolumn{2}{|c|}{Gesture vocabulary}                        \\ \hline
Gesture index & Gesture description                             \\ \hline
G1            & Reaching for needle with right hand             \\ \hline
G2            & Positioning needle                              \\ \hline
G3            & Pushing needle through tissue                   \\ \hline
G4            & Transferring needle from left to right          \\ \hline
G5            & Moving to center with needle in grip            \\ \hline
G6            & Pulling suture with left hand                   \\ \hline
G7            & Pulling suture with right hand                  \\ \hline
G8            & Orienting needle                                \\ \hline
G9            & Using right hand to help tighten suture         \\ \hline
G10           & Loosening more suture                           \\ \hline
G11           & Dropping suture at end and moving to end points \\ \hline
G12           & Reaching for needle with left hand              \\ \hline
G13           & Making C loop around right hand                 \\ \hline
G14           & Reaching for suture with right hand             \\ \hline
G15           & Pulling suture with both hands                  \\ \hline
\end{tabular}
\end{adjustbox}
\caption{ The gestures form a common vocabulary for small action segments that reoccur in different tasks. }
\label{table:gestures}
\end{table}

We first clipped task videos into gesture clips using the annotation files provided with the dataset. We have converted the frame information to time and clipped the tasks into gesture segments. Then we extracted $8$ frames per second from these videos. We resized these frames to $640x480$. We computed the optical flow information of the clipped videos with the method suggested by Thomas \textit{et.al} \cite{flow} and transformed this information into RGB representation images. Although there are $15$ gestures in the gesture vocabulary defined, there is no video data available for the $G7$ : \textit{Pulling Suture with right hand}, so we only have $14$ gesture action labels.  Most studies on JIGSAWS dataset use the gestures through $G1$ to $G11$ only \cite{colin,jigsaws_benchmark}, resulting in $10$ gestures, however, we experiment on all of the gesture labels available.  We rename our data of 14 gestures excluding the empty set of $G7$. For the remaining of this paper we will address them as $<G1,G2,...,G14>$. 

\section{Material and Methods}

Neural networks typically assume that all inputs are independent of each other. However, with sequential tasks and time-series, in order to make a better prediction, we should consider the previous computations and temporal dynamics between the sequential inputs. Recurrent Neural Networks \cite{rnn}, feed-forward networks that unroll over time where each time step is equivalent to a layer, are able to make use of sequential information. In these networks, the hidden unit activations of former inputs of a sequence feeds into the network along with the inputs, making it possible for these networks to learn sequential, time-series models and temporal dynamics. In RNNs, learning is done through back-propagation on the unfolded network. This is called Back Propagation Through Time (BPTT). The total cost function then, is the sum of the error function over time, and at each time step, weights are adjusted according to this error rate. As with feed-forward neural networks, vanishing gradients becomes a problem, and naturally, for the Recurrent Neural Network, this problem leads to exponentially small gradients through time. Recurrent neural networks are difficult to train for this reason. In our work, we specifically use Long Short Term Memory Networks (LSTM) \cite{lstm}, a type of a Recurrent Neural Networks. LSTMs bring a solution to the vanishing gradients problem by introducing the concept of a \textit{memory unit}. LSTMs have a gate mechanism that decides when to forget and when to remember hidden states for future time steps. With this feature, LSTMs are able to train models on long term dependencies.

LRCN introduced by Donahue \textit{et al.} \cite{lrcn} proposes a model that incorporates a deep hierarchical visual feature extractor and a model that can learn long-term temporal dynamics. In this model, a deep hierarchical visual feature extractor, specifically, a Convolutional Neural Network (CNN), takes each visual input $x_{t}$ and extracts its features. This feature transformation of CNN activations produces a fixed length vector representation of the visual inputs:  $\phi V(x_{t})$ which are then passed into a recurrent learning module based on an LSTM. This recurrent model maps inputs to a sequence of hidden states $h_{t}$ and outputs $y_{t}$. In order to predict the distribution at time step {t}, we obtain a vector of class probabilities across our vocabulary by applying softmax to a linear prediction layer on the outputs of the sequential model. 

In the LRCN model for activity recognition, the sequential visual inputs $<x_{1},x_{2},...,x_{T}>$ are first individually processed with Convolutional Networks (CNN), convolutional layer activations are then fed into a single-layer LSTM with 256 hidden units that learns the frames\textsc{\char13} temporal dynamics. Predictions based on averaging scores across all frames from a fixed vocabulary of actions are made. In their study, Donahue \textit{et al.} \cite{lrcn} train two separate LRCN networks on RGB frames and RGB representations of the flow information in the same frames. Decisions are made at a final stage where the network predictions are averaged with fixed weights. 

\subsection{Our Model}

We propose a novel architecture that is based on the principles of LRCN, however our arcchitecture supports multimodal learning and jointly learns temporal dynamics on rich representations of visual features and motion cues. We use the two modalities of the video as input:  RGB frames and the RGB representation of the optical flow information in the same frames. We extract higher level features of the individual frames using CNNs that we train on separate tasks. After we convolve the two streams of input pairs, we concatenate the CNN features and use it as an input to our recurrent model that learns temporal dynamics of consequent frames. Our architecture simultaneously classifies the common low-level gestures and the surgical tasks with a multi-task learning approach. Please see Figure \ref{fig:1} for an overview of our model.  

We draw our input data of RGB and RGB optical flow representations of the same frames from a joint data layer and convolve both these inputs in parallel convolutional neural network (CNN) streams. We concatenate the activations after the 5th convolutional layers. Before pooling this fusion of representations, we apply another convolutional layer and decrease the dimension of the representations. We then apply a fully convolutional layer and feed its output to the LSTM layer along with the sequence clip markers. In our model, we define two different tasks and losses, one for the $3$ surgical task labels which are \textit{Knot Tying}, \textit{Suturing} and \textit{Needle Passing} and another for the 14 different common low-level gesture labels $<G1, G2, G3,...G14>$.

The convolutional body of our architecture is similar to the architecture proposed by Zeiler and Fergus \cite{ZF} and AlexNet \cite{krizhevsky} by Krizhevsky \textit{et al.}, with $5$ convolutional layers, then another one that comes after fusing the activations of these convolutional layers, followed by a fully convolutional layer. We first train two CNNs for a maximum of $40k$ iterations, one for each task; on RGB and flow, that classifies the visual inputs based on individual frames without the notion of temporal dynamics and recurrent networks. We initialize these networks by transferring weights from a pre-trained model on the 1.2M image ILSVRC-2012 dataset \cite{Imagenetsub} for a strong initialization and also to prevent overfitting. We then train two LSTM networks, one for each task; on RGB and flow, with the weights transferred from the individual frame models. We train the flow LSTM and RGB LTSM networks for a maximum iterations of $60k$. Finally, we define our joint model, combining the weights from the convolutional and fully connected layers of the two separate LSTM models we have trained, and transferring them to our model. We perform stochastic gradient descent optimization. We set our learning rate at $0.001$, weight decay at $0.005$, and our learning policy to dynamically decrease the learning rate by a factor of $0.1$ for every $20k$ iterations, for all training models. We train the LSTM network with a maximum iteration number of $90k$. Additionally, we set a threshold (15) based clipping gradient for the LSTM models. The weight for both tasks are equal, so they both contribute to the prediction equally. With time and resource concerns, we chose all mentioned hyperparameters based on our experimental observations with a base of earlier works of Zeiler and Fergus \cite{ZF}, Krizhevsky \cite{krizhevsky} and Donahue \textit{et al.} \cite{lrcn}. Using grid-search, further optimization of these hyperparameters could be possible.

We train our joint model end-to-end with video clips of length greater than $8$ frames. We augment our data by clipping, mirroring, and cropping the frames in order to prevent overfitting.  At each time step, our model predicts both the common low-level gesture and surgical task labels across our vocabulary of gestures and tasks. In order to classify a whole segment of video, 
we average the predictions of each consequent frame for each task. The averaging is done to agree on a single label for the whole video segment. To elaborate, we test our trained model on the multiple clips of $8$ frame length that we extract with a stride of $4$ frames from each video. In order to label the whole video segment, we average the predictions of individual frames and then we average across all clips extracted from the video segment. 

\section{Experiments and Evaluation}

We  evaluate  our novel architecture on JIGSAWS dataset customized to our problem as discussed in the {Dataset} section. We create 6 splits for both the individual frame and LSTM models. We first create randomized lists of videos for training and testing the LSTM models, and then we extract the frames of these videos and randomize them again for the individual frame models. We train our model on a fixed random set of $1200$ gesture video segments and use the rest $422$ for testing. This results in around $42,000$ gesture frames sampled for training and $14,500$ for testing. The video segments have varying sizes from greater than just $8$ frames to $535$, however, most range around $10$ to $90$. We have set our clip frame length on the base of the shorter videos, however, there is room for improvement for this selection. During training we resize both RGB and flow frames to $240x320$ and we augment the data by taking $227x227$ crops and mirroring. We use a setting of multiple $2.62$GHz CPU processors and Geforce GTX$1080$ with computation capability of $6.1$. 

Our experimental results show that our multi-modal and multi-task approach is superior compared to an architecture that classifies the surgical tasks and the common low-level gestures separately on visual cues and motion cues respectively (Table \ref{tab:result}). While the conventional approach reaches a Mean Average Precision (MAP) of only 29.13\% , our architecture reaches a MAP of 50.83\% for 3 tasks and 14 possible gesture labels, resulting in an improvement of 22\% (21.7\%). We observed robust improvements of a median of 22.5\%, ranging from 18\% to 25.1\% for all six split experiments.

% \usepackage{booktabs}
\begin{table}[]
\centering
\caption{For a split of six experiment sets, accuracy for both the single-task, single-modality model and our proposed model are shown. A video segment is defined as accurate only and only when both the gesture and surgical task labels are predicted correctly.}
\label{tab:result}
\begin{tabular}{@{}llll@{}}
\toprule
Split Set               & \begin{tabular}[c]{@{}l@{}}Accuracy for single-task, \\ single-modality model\end{tabular} & \begin{tabular}[c]{@{}l@{}}Accuracy for  \\ Joint Learning Model \\ (multi-task, multi-modal)\end{tabular} & Difference       \\ \midrule
\multicolumn{1}{|l|}{1} & \multicolumn{1}{l|}{26\%}                                                                                                                                                                                               & \multicolumn{1}{l|}{47\%}                                                                                                                                                                            & \multicolumn{1}{l|}{21\%} \\ \midrule
\multicolumn{1}{|l|}{2} & \multicolumn{1}{l|}{30\%}                                                                                                                                                                                               & \multicolumn{1}{l|}{54\%}                                                                                                                                                                            & \multicolumn{1}{l|}{24\%} \\ \midrule
\multicolumn{1}{|l|}{3} & \multicolumn{1}{l|}{29\%}                                                                                                                                                                                               & \multicolumn{1}{l|}{47\%}                                                                                                                                                                            & \multicolumn{1}{l|}{18\%} \\ \midrule
\multicolumn{1}{|l|}{4} & \multicolumn{1}{l|}{29\%}                                                                                                                                                                                               & \multicolumn{1}{l|}{47\%}                                                                                                                                                                            & \multicolumn{1}{l|}{18\%} \\ \midrule
\multicolumn{1}{|l|}{5} & \multicolumn{1}{l|}{30\%}                                                                                                                                                                                               & \multicolumn{1}{l|}{56\%}                                                                                                                                                                            & \multicolumn{1}{l|}{26\%} \\ \midrule
\multicolumn{1}{|l|}{6} & \multicolumn{1}{l|}{30\%}                                                                                                                                                                                               & \multicolumn{1}{l|}{55\%}                                                                                                                                                                            & \multicolumn{1}{l|}{25\%} \\ \midrule
Average                 & 29\%                                                                                                                                                                                                                    & 51\%                                                                                                                                                                                                 & 22\%                      \\ \bottomrule
\end{tabular}
\end{table}

In training, every $5000k$ iterations take about 2 hours and 10 minutes. For a video of frame size ~30, the test time is around 7-8 seconds (excluding the preprocessing time required to extract optical flow, which takes a few seconds for a generic image as recorded by Thomas \textit{et al.} \cite{flow}). Our code does not take full advantage of the GPU and parallelization. For a more efficient study, our code could be modified to fully take advantage of the GPU and parallel computation. 

JIGSAWS dataset benchmark studies \cite{jigsaws_benchmark,colin} train and test on gestures that only belong to one type of surgical task. They train on a set of gestures that occur only during a specific surgical task (e.g. \textit{Suturing}) and also test on this same task. Their experimentation is limited as it trains models only on the gestures that take place in a specific task and not the whole gesture vocabulary of $14$ gestures. Their \textit{Suturing} experiments are done on $10$ gesture labels, while \textit{Knot Tying} and \textit{Needle Passing} are done on only $4$ gestures. Our approach differs greatly; we focus on recognizing a specific gesture whether it takes place in \textit{Suturing} or \textit{Knot Tying}. Therefore, our experiments are done on the whole dataset, across multiple surgical tasks and recognizes all $14$ common low-level gesture labels in our gesture vocabulary.

\section{Conclusion}

We propose  a novel architecture as a solution to simultaneously classify common low-level gestures and surgical tasks in Robot-assisted Surgery (RAS) video segments. Our architecture simultaneously classifies the gestures and surgical tasks with a multi-task learning approach by making use of joint relationships, combining shared and task-specific representations. Our architecture is multimodal and uses two modalities of the video as input: RGB frames and the RGB representation of the optical flow information in the same frames referring to motion cues. After we convolve the two streams of input pairs, we concatenate the CNN features that we extract and use this fusion of higher level representations as an input to our recurrent model that learns temporal dynamics of the gestures and surgical tasks. We focus on recognizing common low-level gestures even when they occur across different tasks. In this manner, it differs greatly compared to the benchmark studies on JIGSAWS \cite{jigsaws_benchmark,colin}. Our multi-modal and multi-task learning approach is superior compared to an architecture that classifies the tasks and the gestures separately on visual cues and motion cues respectively. \\

\textbf{Compliance with ethical standards} \\
\textbf{Conflict of interest} The authors declare that they have no conflict of
interest. \\
\textbf{Ethical approval} This article does not contain any studies with human participants or animals performed by any of the authors.  \\
\textbf{Informed consent} This articles does not contain patient data.

%\begin{acknowledgements}
%\end{acknowledgements}



\end{document}